# Identifying Planetary Names in Astronomy Papers: A Multi-Step Approach

Golnaz Shapurian, Michael J Kurtz, and Alberto Accomazzi

Abstract

The automatic identification of planetary feature names in astronomy publications presents numerous challenges. These features include craters, defined as roughly circular depressions resulting from impact or volcanic activity; dorsas, which are elongate raised structures or wrinkle ridges; and lacus, small irregular patches of dark, smooth material on the Moon, referred to as "lake" (Planetary Names Working Group, n.d.). Many feature names overlap with places or people's names that they are named after, for example, Syria, Tempe, Einstein, and Sagan, to name a few (U.S. Geological Survey, n.d.). Some feature names have been used in many contexts, for instance, Apollo, which can refer to mission, program, sample, astronaut, seismic, seismometers, core, era, data, collection, instrument, and station, in addition to the crater on the Moon. Some feature names can appear in the text as adjectives, like the lunar craters Black, Green, and White. Some feature names in other contexts serve as directions, like craters West and South on the Moon. Additionally, some features share identical names across different celestial bodies, requiring disambiguation, such as the Adams crater, which exists on both the Moon and Mars. We present a multi-step pipeline combining rule-based filtering, statistical relevance analysis, part-of-speech (POS) tagging, named entity recognition (NER) model, hybrid keyword harvesting, knowledge graph (KG) matching, and inference with a locally installed large language model (LLM) to reliably identify planetary names despite these challenges. When evaluated on a dataset of astronomy papers from the Astrophysics Data System (ADS), this methodology achieves an F1-score over 0.97 in disambiguating planetary feature names.

1. Introduction

Identifying mentions of planetary surface features such as craters, dorsa, and lacus in astronomy literature presents considerable challenges. Planetary feature names exhibit significant lexical ambiguity; they overlap with common words, personal names, and locations that features are named after. For instance, crater names like Newton or Copernicus match renowned scientists. Additionally, many identical feature names are shared across celestial bodies. For example, craters named Kuiper exist on three separate celestial bodies: the Moon, Mars, and Mercury. Reliably recognizing ambiguous features requires disambiguation based on a wider context.

Standard named entity recognition (NER) models struggle with this task, as feature name mentions appear in similar local syntax referring to different celestial body surfaces. With limited contextual differentiation within a sentence, NER models cannot easily distinguish identical names on different planets and moons.

To address these challenges, we have developed a multi-step pipeline combining rule-based candidate retrieval, statistical relevance analysis, part-of-speech (POS) tagging, NER filtering, hybrid keyword harvesting, knowledge graph (KG) matching, paper relevance score, and interaction with a locally installed large language model (LLM).

Together, these techniques enable disambiguating and validating planetary feature mentions despite significant ambiguity. Customized rules retrieve candidate names, while statistical relevance analysis provides contextual clues to differentiate identical names by parent body. The NLP techniques and KGs address sparse contextual cues. Additionally, the paper relevance scoring and LLM analysis incorporate global signals that complement the context-focused methods.

This comprehensive methodology overcomes lexical, semantic, and contextual ambiguity to accurately identify challenging planetary names in astronomy literature. The hybrid techniques provide robustness lacking in standard named entity recognition models applied alone.

2. Planetary Nomenclature

Currently, there are over 16,000 officially approved names referring to various surface features across celestial bodies in our solar system (U.S. Geological Survey, n.d.). Each name links to a specific target planet or moon and a feature type like a satellite feature, crater, or albedo feature. For instance, the feature name Arabia refers specifically to an albedo feature on Mars.

Satellite features have the highest frequency of assigned names across celestial bodies, with over 7,100 uniquely named instances approved thus far. The Moon contains over 9,000 named features in total, with craters representing the most common type at over 1,600 named entities. Craters also make up the majority of named features on Mars, with over 1,100 distinctly named craters. Interestingly, over 130 craters share identical names on both celestial bodies, including Galle, McLaughlin, and Isis, which exist as named craters on the surfaces of the Moon and Mars.

With thousands of named features across just the Moon and Mars, substantial ambiguity arises. For example, Mare Australe is a large dark plain on the Moon, but it is also the name of an albedo feature on Mars. The expansive and growing lexicon poses considerable challenges for automated feature extraction.

3. Related Work

Automating the construction of a structured KG from scientific papers (Mondal et al., 2021), involved extracting key entities such as tasks, datasets, and metrics, which were then treated as nodes. These entities were interconnected by relations such as coreference, with cross-document coreference links employed to canonicalize entity mentions. This innovative methodology demonstrates the utility of KGs and entity linking for literature analysis across documents. Similarly, we have implemented an approach utilizing cross-document coreference to apply a KG for disambiguating identical planetary feature names appearing across different celestial bodies.

In a recent study (Chen et al., 2023), researchers explored leveraging LLMs for NER in astronomical literature without annotated training data. Their Prompt-NER approach adapted models like GPT-3.5, LLaMA, and Claude, utilizing task descriptions, entity definitions, emphasis prompts, and examples. Experiments by Chen et al. on annotated astronomy articles and telegram text demonstrated significant performance improvement through prompt engineering, achieving F1-scores exceeding 0.75 on articles and over 0.90 on telegrams. This study underscores prompt-based techniques as a promising approach for rapidly adapting LLMs to entity extraction in scientific texts without field-specific training data.

Our hybrid pipeline uniquely combines NLP, NER, and KG techniques with an LLM for robustness. We have specifically addressed lexical ambiguity, considering not only identical planetary feature names appearing across different celestial bodies but also entities with varying contextual information beyond the planetary names. This comprehensive methodology fuses complementary global and local contextual signals to provide disambiguation capabilities and flexibility, aspects lacking in approaches that solely rely on LLMs or KGs applied in isolation. Our system achieves an F1-score of over 0.97 in disambiguating planetary feature names.

Although computationally intensive, our approach prioritizes quickly narrowing down results in the earlier stages of the pipeline. This allows later stages to invest more computation on fewer results, including interacting with the locally installed LLM via prompts. This interaction is essential for the effective validation of filtered candidate names.

4. Pipeline

In this section, we will delve into the specifics of our multi-step pipeline for identifying planetary names in astronomy papers. This pipeline is designed to address the challenges posed by the variability in planetary name usage, the lack of context, and the potential for false positives.

To reliably and accurately extract planetary names from full text, the proposed pipeline employs a combination of techniques. These include rule-based search querying to retrieve candidate feature names, NER filtering to eliminate false positives not referring to planetary surfaces, POS tagging to analyze local context, KG matching to model semantic relationships, and calculating a probability score to indicate the likelihood of valid planetary names based on key text characteristics. The pipeline also includes querying a locally installed LLM to incorporate global document understanding. The pipeline stages are shown in Figure 1.

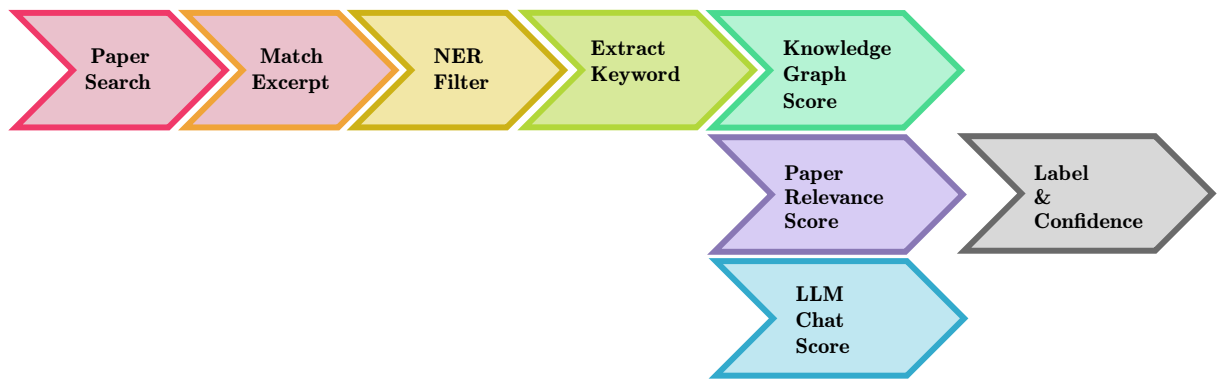

Figure 1: Pipeline for extracting planetary feature names from full text. The pipeline consists of candidate retrieval, false positive filtering, context analysis, KG matching, paper analysis, and language model querying stages.

The following subsections discuss each pipeline stage in detail, highlighting their role and how they combine to robustly identify challenging feature mentions.

4.1. Paper Search and Retrieval

The first step of the pipeline searches the Astrophysics Data System (ADS) using application programming interface (Kurtz et al., 2000) to retrieve candidate feature names appearing alongside the target planetary name. This identifies potential feature names that may be relevant to the planetary name and to reduce the number of false positives. The search process uses a filtering algorithm that takes into account the context of the feature names, such as the distance between the feature name and the target planetary name, to prioritize the most relevant features.

4.2. Extracting Matched Excerpt Windows

After retrieving candidate records, the next key step is determining if the feature name refers specifically to the celestial body being analyzed. This disambiguation is essential since some feature names like Herschel exist on multiple celestial objects—in this case, as craters on the Moon, Mars, and Saturn's moon Mimas.

To ascertain relevance, a keyword analysis technique is utilized. This involves identifying keywords associated with each celestial body and comparing their frequencies in the text. For example, a higher frequency of keywords like "Moon", "Luna", or "Lunar" suggests the feature name Herschel refers to the lunar crater Herschel. Similarly, a higher occurrence of terms like "Mars" or "Martian" implies it is the Martian crater, Herschel. And a greater presence of words such as "Mimas" or "Saturnian" indicates it is the Herschel crater on Mimas. The probabilities are calculated by tallying the contextual keywords, determining their relative percentages, and interpreting the scores.

If one target has a distinctly high probability that it does not match the celestial body being analyzed, the record is filtered out as irrelevant. However, if the probabilities between multiple bodies are within a specified range or the high score matches the body being analyzed, the record is retained for further processing. Specifically, records are kept if their probability score is greater than or equal to $1/N$, where N represents the total number of celestial bodies included in the analysis. For instance, when comparing 3 bodies like Mars, the Moon, and Mimas, records with probabilities exceeding 33.3 percentage points are passed along for further analysis. This threshold ensures that cases with genuinely ambiguous celestial body connections or correct match alignment are passed along for subsequent consideration instead of being prematurely discarded.

After identifying relevant records, a regular expression is used to identify the feature names in the text, and a window of 129 tokens around the feature name is identified for further analysis.

Next, SpaCy's (spaCy, n.d.) POS tagging is utilized to determine whether the feature name has appeared in the context as an adjective. For example, consider the feature name Black (a crater on Moon) in the following sentence: "Black indicates that a crater is fresh; light gray means that a crater is highly degraded from diffusive degradation processes." Since "Black" serves as an adjective describing the freshness of the crater, this excerpt will be eliminated from consideration, which helps to diminish the number of false positives and increase the accuracy of the feature name identification process.

Additionally, the algorithm examines extracted noun phrases to ensure the feature name is not contained within larger expressions. For instance, if the planetary name to identify is Alamos or Circle (both craters on Mars), the algorithm excludes matched excerpts containing Los Alamos (a location in New Mexico) or Arctic Circle (a geographical feature). Similarly, if the planetary name to identify is Qidu (crater on Mars) or Siddons (crater on Venus), the algorithm eliminates matched segments having Qidu Fossae (Fossa on Mars) or Siddons Patera (Patera on Venus).

4.3. Leveraging Named Entity Recognition Model

The next step in the pipeline is to remove planetary names that have different meanings in their context. This is achieved by utilizing AstroBERT NER (Grezes et al., 2021), a powerful tool that can recognize and classify named entities in astronomy-related text. AstroBERT NER has been fine-tuned on a large dataset of scientific astronomy papers, enabling it to distinguish between various types of celestial objects, organizations, and observatories. Additionally, it can recognize citations, people, grants, and fellowships.

For example, the names "Walker", "Wallace", "Walter", "Watson", "Wilder", "Wilhelm", "Williams", and "Wilson" all refer to craters on the Moon, Mars, and Venus. However, these names may also appear in articles as references, people, grants, or fellowships rather than planetary features. AstroBERT NER accurately identifies these instances, and hence they are eliminated, ensuring that only instances of feature names used in the context of planetary surfaces are included in the next stage of pipeline processing.

In addition to citation, person, grant, and fellowship entities, AstroBERT NER identifies and classifies other types of entities in astronomy-related text, such as locations, telescopes, missions, and models. For instance, Aspen and Philadelphia, which are large cities in the US, are also craters on Mars, while Tempe in Arizona is an albedo feature on Mars. By categorizing these entities as locations, AstroBERT NER flags them for elimination from the pipeline, removing false positive matches from further processing.

Similarly, Hubble (Hubble Space Telescope), Plato (Plato's Reflecting Telescope), and Webb (James Webb Space Telescope) are craters on the Moon and have been referred to in astronomy papers as telescopes. Apollo (Apollo Lunar Landing Mission) is a crater on the Moon, while Cassini (Cassini-Huygens Mission to Saturn and its Moons) and Copernicus (Copernicus Planetary Mission to the Moon) are craters on the Moon and Mars. Lambert (Lambert's Reflectance Model) is a crater on both the Moon and Mars, while Euler (Euler equation models) and Bingham (Bingham Plastic Models) are craters on the Moon. When these entities appear in references to something other than the planetary surface, AstroBERT NER identifies them to be eliminated, reducing the risk of false positives in identifying planetary names.

By identifying feature names as celestial objects or regions, or not identifying them at all, AstroBERT NER filters matched entities for further processing in the pipeline.

4.4. Top Keyword Extraction

In this step of the pipeline, the top keywords are extracted from the excerpt window surrounding each candidate feature mention. These keywords help determine if the name appears in a planetary or non-planetary context. Three methods are used for keyword extraction.

Yake (LIAAD, n.d.) is a Python library that extracts keywords using frequency analysis and deduplication thresholds. It is applied to identify 2-word n-grams in the excerpts using settings tuned for this domain. Yake determines the top 20 keywords in each excerpt by evaluating their significance and relevance within the context of the text, producing a list of the most crucial terms.

SpaCy's NER module, part of their pretrained English pipeline, is utilized to extract keyword entities from the text. SpaCy NER can identify locations, organizations, people, products, and general nouns, among other categories. Only keywords labeled as one of these entity types are kept for further analysis. By extracting only nouns categorized as semantic entities, SpaCy NER filters out irrelevant verbs, adjectives, and other parts of speech while capturing keywords meaningful to the context. The identified locations, organizations, persons, products, and nouns serve as useful signals for disambiguating planetary feature mentions based on the surrounding entities.

A custom Wikidata (Wikidata, n.d.) vocabulary is created by extracting planetary names from Wikidata, encompassing a wide range of astronomical concepts and phenomena. These terms include the names of celestial bodies such as stars (e.g., Sirius), planets (e.g., Jupiter), moons (e.g., Europa), asteroids (e.g., Ceres), comets (e.g., Halley's), and deep-sky objects (e.g., Crab Nebula, Andromeda Galaxy). Additionally, the list includes names of astronomical phenomena like variable stars, novas, supernovas, pulsars, bursts, and binary systems. The keywords also cover surface features on planetary bodies, such as mountains (e.g., Olympus Mons), valleys (e.g., Valles Marineris), and craters (e.g., Basin). Furthermore, the extracted terms include important scientific concepts related to planetary astronomy, such as surface formations, ring systems, planet types based on motion, orbits around planets, and positions relative to planets. Examples of such terms include quadrangle, ring, inferior/superior planet, retrograde motion, circumplanetary, synchronous orbits, and sub-Earth point.

This domain-specific astronomical terminology complements the more general keywords produced by Yake and SpaCy. Without these extra terms, Yake and SpaCy may fail to reliably recognize crucial astronomy concepts due to a lack of built-in knowledge in this field. By supplementing their outputs with Wikidata's structured terminology, the model gains the capacity to better disambiguate valid planetary feature mentions based on the presence of these domain terms in the excerpt.

The broad terminology extracted from Wikidata provides a benchmark for comparing the keywords present in each excerpt. The presence of these planetary terms in the excerpt signals that the context is discussing a valid planetary feature.

After extracting keywords from each excerpt using Yake, SpaCy, and Wikidata, the terms are lemmatized to consolidate different morphological forms. Identical keywords between Yake and SpaCy are merged and added to the list first. Next, the unique Wikidata terms are appended.

To reach a total of 10 keywords per excerpt, additional terms are drawn alternately from the Yake and SpaCy outputs as needed. This creates a consolidated set of the top 10 lemmatized keywords, representing the dominant context of each excerpt.

4.5. Disambiguation with Knowledge Graphs

Similar to (Mondal et al., 2021), we adapt the KG approach for disambiguating planetary feature names mentioned in astronomical publications. A KG (Stanford AI Lab, 2019) is a structured representation of concepts and their relationships, comprising nodes, edges, and weights. A node represents a distinct entity, in this case, feature names, feature types, and top keywords.

An edge interconnects nodes and represents the relationships between concepts, such as "part-of" or "appeared-with." For example, the node for Apollo, a feature name, has a "part-of" relationship with the node for craters, a feature type. Or the top keywords appearing with Apollo in an excerpt, such as "depression", "upper limit", or "light plains", have an "appeared-with" relationship with Apollo.

The weights denote how often two entities co-occur. The higher the weight of an edge, the stronger the implied bond between the entities. This graph structure enables an efficient lookup of a concept's semantic context based on its neighboring nodes and edges. Traversing local neighborhoods provides contextual clues.

Consider the example of a KG with a node for Apollo, representing a crater. To disambiguate another crater on the Moon, like Balmer, looking at the connected nodes and finding an edge to "light plains" would indicate Balmer may also be a crater in that excerpt. The more nodes the top keywords of Balmer share with recognized craters, and the higher those connections' weights, the higher the confidence Balmer is a crater in that excerpt.

In this way, KGs can disambiguate feature names by examining connections between surrounding keyword nodes and conceptual relationships. The graph structure efficiently incorporates domain knowledge into disambiguation.

The pipeline creates two KGs in this step for each feature name: one where the name appears in the context of a planetary surface and one for other contexts. To build the graphs, a collection of labeled excerpts containing the feature name is processed to extract keywords. One KG incorporates keywords from excerpts referring to planetary surfaces. The other graph contains keywords from excerpts where the name refers to something else.

When predicting whether a feature name in an unknown excerpt is referencing a planetary surface, its keywords are identified and queried against both KGs. This produces a probability score indicating if the name refers to a planetary surface feature or not.

The probability score is calculated by summing the total edge weights for all paths found for all keywords in each KG. Specifically, for the planetary surface KG, the edge weights of all paths connecting every excerpt keyword to the feature name node are totaled, and a similar calculation is carried out for the non-planetary entity context KG. The sum of these edge weights from the planetary surface KG is then divided by the total sum of weights from both graphs to determine the final probability, indicating the likelihood that the feature name has been employed in the context of a planetary surface.

The hypothesis posits that each feature name is associated with a distinct vocabulary. This notion is visually exemplified by the Antonaidi Moon and Mars craters, as depicted in the word cloud Figure 2. A corpus of 72 papers pertaining to the Antonaidi Moon crater and 56 papers regarding the Antonaidi Mars crater are selected for training. The excerpts from these texts contain 637 and 447 unique keywords, respectively. Notably, only 28 keywords are shared between the two corpora.

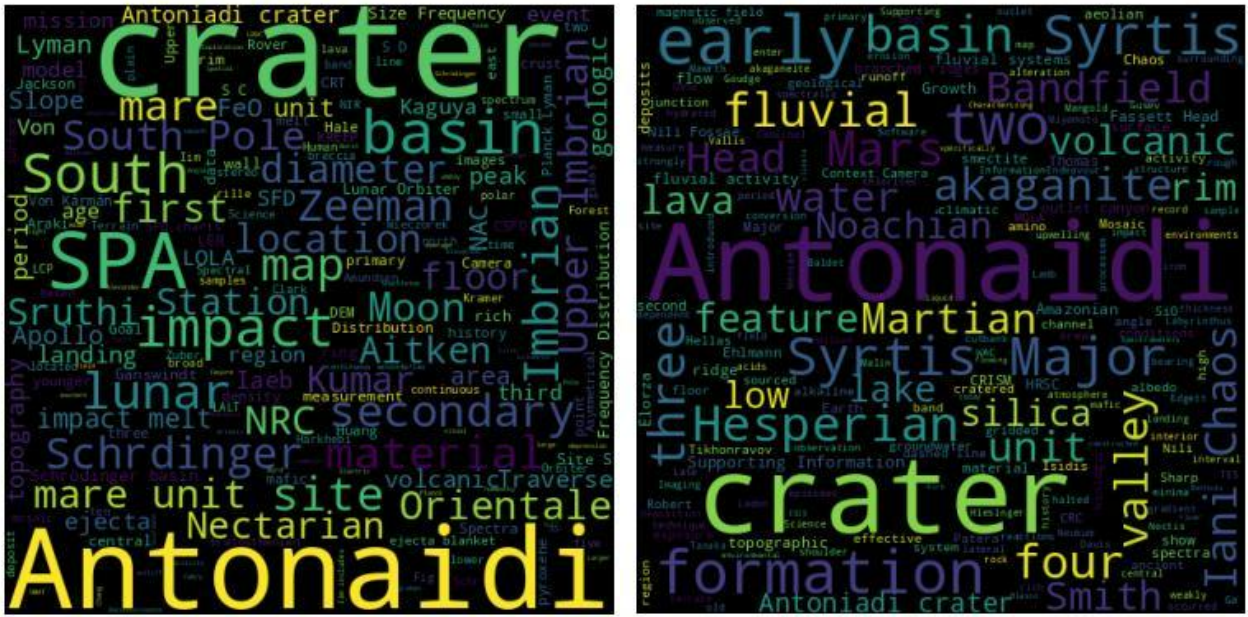

Figure 2: Visualizing Feature Name Vocabulary Distinctions - Antonaidi Moon and Mars Craters

## 4.6. Paper Relevance Score

A probability score indicating the likelihood that valid planetary feature names appear in the paper is calculated based on five key characteristics. This score is necessary to provide additional context beyond the local excerpt keywords when determining if a candidate name truly refers to a planetary surface feature. The properties used to compute this probability are: whether the paper appeared in an astronomy collection in ADS, if it was published in a planetary science journal, if the target celestial body (e.g., Moon, Mars) appeared more than a threshold number of times, if the feature type (e.g., crater, satellite feature) was mentioned more than a threshold number of times, and whether at least a threshold number of Wikidata-derived planetary astronomy keywords were used overall. Papers published in specialized astronomy publications and journals focused on planetary bodies, which also sufficiently reference those targets and feature types using ample domain terminology, have a higher probability of containing valid planetary feature name mentions. Weighing these signals helps disambiguate names in papers that lack strong local contextual cues within each candidate excerpt. The aggregated paper-level clues complement the narrow excerpt-based features to better estimate if a name refers to a planetary surface feature.

## 4.7. Leveraging Large Language Models

Recent work (Chen et al., 2023) demonstrated the potential of leveraging large language models (LLMs) for named entity recognition in astronomical texts without needing annotated training data. Motivated by their success using prompt engineering to achieve high performance, we incorporated Orca Mini (Mathur, 2023), a freely available LLM, into our pipeline to further improve the disambiguation of lexically ambiguous planetary feature names.

For each candidate excerpt, Orca Mini is provided with the paper's title and abstract in a broader context and asked to judge whether the feature name in the excerpt refers to a planetary surface feature or not. It answers yes or no, with a score of 1 for yes and 0 for no, indicating its confidence in that prediction based on comprehending the larger paper context. We included this LLM stage because names that are ambiguous in their local excerpt may be clarified when considering the overall paper topic and content. The LLM's understanding of the paper can provide additional signals to determine if a feature name truly refers to a planetary surface, complementing the pipeline steps focused on analyzing each excerpt in isolation. By incorporating both local excerpt-level and global paper-level context, Orca Mini's predictions help resolve remaining ambiguities and improve the accuracy of feature name identification.

## 4.8. Label and Confidence Score

Each instance of an identified feature name is given a label indicating whether it has appeared in the context of a planetary surface or not, as well as a confidence value representing the accuracy of the label. The label is identified based on three scores: KG score, paper relevance score, and LLM chat score. These three scores are fed into a support vector machine (Vapnik et al., 1995) model that was trained on over 20,000 verified datapoints to predict the label and confidence score.

## 5. Results

Two experiments were conducted to evaluate the performance of the pipeline stages by analyzing how different feature names are processed across multiple disambiguation steps. One analysis examined craters with identical names on both the Moon and Mars, while another examined craters with names used in multiple other contexts.

There are about 140 feature names that have been used multiple times for planetary names, of which 132 are craters for both the Moon and Mars, including Kaiser. The search for Kaiser crater returned 52 records related to the lunar crater and 77 pertaining to the Martian crater. Of these, 33 records were included in the results for both searches.

As previously discussed, filters are applied through various stages of the pipeline to further refine the results. One record was excluded for being in a non-English language.

Twelve records were omitted for not mentioning the Moon or Mars in the full text. For instance, the record "A CO2 Cycle on Ariel? Radiolytic Production and Migration to Low-latitude Cold Traps" (Cartwright, 2022) was initially included in the Mars search due to an author's address containing "Mars Hill Road." However, the pipeline correctly excluded this record from the Mars search results since the main text did not mention Mars. The same record was also included in the Moon search, but because the text only referred to Uranus's moons using lowercase "moon," the pipeline eliminated it from the Moon search as well.

Thirty-one additional records were considered irrelevant to celestial bodies and were filtered out. For instance, the record "Active Mars: A Dynamic World" (Dundas et al., 2021) appeared in both the Mars and Moon searches. The match excerpt stage computes likelihood percentages to indicate which celestial body the shared record refers to, and with a 99.29% probability, it concluded that the excerpt refers to the Mars crater Kaiser.

Forty records were filtered out using NLP processing techniques, which identified Kaiser as part of phrases. For instance, the record "Spectroscopic study of perchlorates and other oxygen chlorides in a Martian environmental chamber" (Wu et al., 2016) contains "Kaiser Optical Systems Inc".

Twenty records were excluded by AstroBERT NER. For instance, the record "A search for Vulcanoids with the STEREO Heliospheric Imager" (Jones et al., 2022), included in both searches, was omitted due to AstroBERT NER identifying the only instance of Kaiser as a citation.

One record was eliminated due to the regular expression not matching the feature name in the excerpt (e.g., if the feature name appears in lowercase or hyphenated form). The record, "Astrophysics in 2006" (Trimble et al., 2007) was eliminated because its sole instance of Kaiser is "Kaiser-Frazer".

After applying all filters, the final set contained 20 records specific to the Martian Kaiser crater that were suitable for analysis, and none related to the lunar Kaiser crater.

The 20 Mars-related records contained multiple references to the Kaiser crater feature name. After excerpting all mentions of Kaiser crater from these records, there were a total of 54. All are correctly labeled as pertaining to the Kaiser crater on Mars.

The following two excerpts are from "Arecibo radar imagery of Mars: II. Chryse-Xanthe, polar caps, and other regions," (Harmon et al., 2017):

*Surrounding these two dark craters is radar-bright terrain corresponding to highly eroded surfaces in and around the valley networks. To the south can be seen two bright-rim crater features associated with a crater at 11.8E, 40.1S (16) and Kaiser Crater (17). This first of these is a complete ring feature that corresponds to a highly terraced and eroded interior rim such as seen for the bright-rim craters in Arabia.*

*This first of these is a complete ring feature that corresponds to a highly terraced and eroded interior rim such as seen for the bright-rim craters in Arabia. The Kaiser feature ( Figs. 18 and 20 ) is only a partial arc that corresponds to the highly eroded and gullied east and northeast interior rims of this much larger crater. 6 Large basins Three prominent basin-size impact structures (Hellas, Argyre, and Isidis) show up as depolarized radar features.*

The Kaiser example provides a detailed examination of how the pipeline processed an ambiguous feature name through various disambiguation stages. Table 1 summarizes the overall results for all 132 craters sharing designations between the Moon and Mars.

| Crater feature name shared between the Moon and Mars | Moon Records | Mars Records |
|---|---|---|
| Initial search records returned | 20092 | 18342 |
| Shared crater names between Moon and Mars | 132 | 132 |
| Filtered - Non-English language | 0 | 12 |
| Filtered - Irrelevant to celestial body | 7307 | 3957 |
| Filtered - No mention of celestial body in full text | 2655 | 302 |
| Filtered - NLP techniques | 3901 | 5542 |
| Filtered - astroBERT NER | 2490 | 3769 |
| Filtered - Feature name not matched by regular expression | 2619 | 3291 |
| Filtered - Feature name detected in a phrase in keywords | 438 | 537 |
| Records considered for further processing | 682 | 932 |
| Excerpts considered for further processing | 2533 | 3726 |
| Feature name correctly labeled (True Positive) | 1807 | 2450 |
| Feature name incorrectly labeled (False Positive) | 63 | 87 |
| Entity incorrectly labeled (False Negative) | 74 | 105 |
| Precision | 0.96 | 0.96 |
| Recall | 0.97 | 0.97 |
| F1-score | 0.96 | 0.96 |

Table 1: Summary of planetary feature name extraction results for craters shared between the Moon and Mars. The table shows the volume of records retrieved, filtered, and analyzed through each stage of the pipeline, along with precision, recall and F1-score for crater names to their respective celestial bodies.

For the lunar experiments, out of 2,553 identified feature names, 1,807 were correctly recognized as planetary feature names, 63 were incorrect, and 74 were missed, achieving an F1-score of 0.96. For Mars, out of 3,726 identified feature names, 2,450 were accurately identified as planetary feature names, 87 were wrong, and 105 were overlooked, also achieving an F1-score of 0.96.

The table enumerates the initial records retrieved for lunar and Martian instances and the number filtered out at each pipeline stage. Although thousands of initial records may be found, the pipeline narrowed these down to a small relevant subset. Analyzing the shared crater names demonstrates robust capabilities in handling lexical ambiguity and precisely extracting entities despite identical titles across celestial bodies.

There are numerous planetary feature names that share names with prominent scientists, but not all of them are named after the scientists themselves. For instance, the crater Hill is named after George William Hill, while the Lambert and Euler craters are not named after Johann Heinrich Lambert and Leonhard Euler, respectively. However, the similarity in names can create ambiguity when their surnames appear in astronomical literature about planetary surfaces and features. This is because Euler, Lambert, and Hill made major contributions to foundational concepts in physics, mathematics, and astronomy that enabled many later advances in planetary science. As such, their names have been extensively used for equations, methods, theorems, stability criteria, and other entities across these disciplines, as shown in the table in the appendix.

When their surnames were encountered in publications related to planetary science, it required careful discernment to determine if the author is referring specifically to the lunar crater Hill named for George William Hill versus the many principles, techniques, and features named after Hill the scientist. The same issue arose for the disambiguating usages of Lambert and Euler. Hence, another analysis was performed to examine lunar craters Euler, Lambert, and Hill, and Martian crater Lambert, to evaluate the accuracy of the pipeline.

After an initial search and filtering, 60 records were selected for further pipeline processing. Among these, the four crater entities were identified in 63 excerpts—labeled as planetary and 39 as non-planetary. All excerpts were verified by an expert as correctly categorized.

The pipeline performance for the four aforementioned craters is summarized in Table 2. For Euler crater, 40 of the 52 excerpts were correctly identified as referring to the planetary name (true positives), with the remaining 12 excerpts correctly eliminated. This yielded an F1-score of 1. The lunar Lambert crater had 7 true positives out of 8 relevant excerpts, with only 1 non-relevant excerpt incorrectly kept. This achieved a 0.93 F1-score. The Martian Lambert saw 11 of 12 relevant excerpts correctly classified as such, giving a 0.96 F1-score. Finally, Hill crater attained 5 true positives out of 5 relevant excerpts, with the remaining non-relevant excerpts correctly excluded for a 1 F1-score. Across all four craters, the high precision and recall translates to strong overall F1-score results, demonstrating reliable distinction between relevant and non-relevant excerpts.

This section demonstrated the capabilities of our pipeline in extracting shared feature names between the Moon and Mars, illustrated by in-depth analysis of the Kaiser crater. The precise identification of these crater names, which exhibit identical nomenclature on both celestial bodies, serves as a robust validation of our pipeline's prowess in disambiguation, labeling, and retrieval techniques. Furthermore, our inspection of the Euler, Lambert lunar, Lambert Martian, and Hill craters underscores the pipeline's precise named entity extraction and disambiguation, particularly when confronted with feature names closely resembling a multitude of other entities.

| Craters | Euler Records | Lambert (Lunar) Records | Lambert (Martian) Records | Hill Records |
|---|---|---|---|---|
| Initial search records returned | 93 | 307 | 309 | 592 |
| Records considered for further processing | 20 | 5 | 19 | 16 |
| Excerpts considered for further processing | 52 | 9 | 21 | 21 |
| Feature name correctly labeled (True Positive) | 40 | 7 | 11 | 5 |
| Feature name incorrectly labeled (False Positive) | 0 | 1 | 1 | 0 |
| Entity correctly labeled (True Negative) | 12 | 1 | 10 | 16 |
| Entity incorrectly labeled (False Negative) | 0 | 0 | 0 | 0 |
| Precision | 1 | 0.88 | 0.92 | 1 |
| Recall | 1 | 1 | 1 | 1 |
| F1-score | 1 | 0.93 | 0.96 | 1 |

Table 2: Breakdown of pipeline labeling results for four crater entities examined. The table shows the number of records processed, excerpts extracted, count of true and false positives, count of false negative, precision, recall and F1-score, for the four crater entities examined.

Table 3 displays the confusion matrix aggregating performance over all crater identification. The pipeline achieved strong precision (0.96), recall (0.99), and F1-score (0.97) for detecting craters. Our strategic emphasis has been on achieving a high recall rate to minimize the risk of overlooking any potential entity, ensuring comprehensive identification. The near-perfect recall of 0.99 attests to the effectiveness of our approach in successfully capturing almost all real craters present in the astronomy papers indexed in ADS. The high precision of 0.96 indicates that most predicted craters are accurate. Meanwhile, the balanced F1-score of 0.97 demonstrates robust crater recognition abilities. This approach aligns with the importance of not missing any potential entities and enhances the model's ability to provide a thorough and reliable planetary feature name recognition.

| Craters | Predicted + | Predicted - |
|---|---|---|
| Actual + | 41519 | 408 |
| Actual - | 1768 | 4703 |

Table 3: Confusion matrix for crater identification.

6. Discussion

In comparing our pipeline to established statistical NER models like BERT (Devlin et al., 2019), Stanford NER (The Stanford NLP Group, n.d.), and SpaCy, we've observed a significant improvement in handling highly ambiguous entities. Although specific quantitative metrics were not collected in this study, the decision to develop a custom system stemmed from unsatisfactory performance with widely used models in initial attempts. The shortcomings of these models in our context led to the creation of our pipeline. This multi-stage approach, integrating NLP and knowledge-based techniques, consistently demonstrates superior efficacy in addressing challenges posed by highly ambiguous entities.

The success of our pipeline lies in its ability to outperform traditional statistical NER models through an intricate process. By integrating NLP and knowledge-based techniques, our multi-stage pipeline enables comprehensive disambiguation. The initial search, POS tagging, and astroBERT NER filter out irrelevant entities, while hybrid keyword harvesting generates descriptive keywords for each entity. The KG models semantic connections between entities and keywords, facilitating informed disambiguation using contextual clues.

At disambiguation time, the keyword set from a text excerpt is compared against the KG to find the best matching entity by semantic similarity. The graph's rich keyword connections allow for the disambiguation of ambiguous references. Furthermore, paper relevance score and having LLM analyze the excerpt and provide a probability score of its relevance offers additional capabilities to handle difficult cases.

The pipeline's disambiguation capabilities are robust, but there is room for improvement, especially in cases where mentions are sparse or ambiguous. To tackle this challenge, additional techniques can be employed, such as clustering textual excerpts that contain ambiguous terms. This approach provides more context and helps differentiate between rarely mentioned entities. By clustering textual excerpts with the same ambiguous term, we can aggregate contextual information across multiple mentions, resulting in a more comprehensive representation of entities. Expanding the KG to include lesser-known entities will also enhance coverage and overall performance. Analyzing larger excerpt windows can further enrich the contextual information. By combining these techniques, we can effectively improve the disambiguation of sparse or obscure cases.

Ultimately, this disambiguation tool will enable more accurate analysis of planetary science texts by extracting precise entity references. The flexible pipeline design allows for expansion as new features are discovered. Our multi-faceted approach represents a significant advancement in named entity disambiguation for planetary science. Ongoing development will enable deeper understanding and knowledge extraction from planetary science text data.

7. Conclusion

Identifying planetary surface features like craters, dorsas, and lacus in astronomy literature presents significant challenges due to the lexical ambiguity of feature names. As discussed, identical names are shared across celestial bodies and can refer to people, places, and other entities. Standard NER techniques fail to disambiguate these lexically ambiguous names using only the local context.

To address the challenges of identifying planetary feature names in astronomy literature, we have developed a multi-stage hybrid pipeline that combines NLP techniques, a NER model, keyword extraction, KGs, and LLM queries. The pipeline filters candidate names using rules, matches local context via keywords and the KG, and determines the overall likelihood of a feature name referring to a planetary surface using global context from the LLM and paper relevance scoring.

Our robust methodology overcomes the limitations of a standard NER model applied in isolation. By incorporating both local context and global semantic knowledge, the system can reliably identify even highly ambiguous planetary feature names. The extraction of these named entities enables the creation of authoritative gazetteers, supporting planetary science and space exploration.

Moving forward, continued enrichment of the KG and model training on new literature can further improve coverage and accuracy. Overall, this hybrid technique represents an important advance in automatically identifying challenging domain entities from text.

Acknowledgements

We would like to thank our colleague Thomas Allen for acting as the domain expert, verifying that the data labeled as either planetary or non-planetary was appropriately categorized during the implementation of the pipeline. Additionally, we greatly appreciate the efforts of Sergi Blanco-Cuaresma for handling the installation and configuration of the Orca Mini language model used in the pipeline stage. Their expertise with data validation and implementation of high-performance computing infrastructure, respectively, allowed us to focus fully on developing the methodology.

Appendix

The discoveries and key contributions of the mathematicians Leonhard Euler, Johann Lambert, and George William Hill across various fields like astronomy/astrophysics, physics/optics, mathematics, and cartography are summarized in the table below.

| **Astronomy / Astrophysics** | * Lunar crater Euler - Lunar crater named after Leonhard Euler, Swiss mathematician (1707-1783). |
|---|---|
| | * Lunar crater Lambert - Crater on the Moon named after Johann Heinrich Lambert, German astronomer, mathematician, physicist (1728-1777). |
| | * Martin crater Lambert - Crater on Mars named after Johann Heinrich Lambert German astronomer, mathematician, physicist (1728-1777).
* Lambert's theorem - Computes orbital velocity of an object based on distance from the central body.
* Lambert parameters - Orbital mechanics parameters used to calculate transfers between orbits around a central body. Specify radii, anomalies, time. |
| | * Lunar crater Hill - Crater on the Moon named after George William Hill, American astronomer, mathematician (1838-1914).
* Hill sphere - The region around a smaller orbiting body where it dominates gravitationally over the larger body it orbits.
* Hill radii - The radii of Hill spheres. Determines the sphere of gravitational influence.
* Hill problem - Studies the motion of a small body under the gravitational influence of two larger orbiting bodies.
* Hill determinant - A function used to assess the stability of solutions to the Hill problem.
* Hill's lunar theory - A theory to explain the motion of the Moon under the gravitation of the Earth and Sun.
* Hill's equation - Describes motion of orbiting bodies subject to perturbations.
* Hill stability - The stability of an orbit perturbed by a second smaller orbiting body.
* Hill stability criterion - Mathematical criteria to determine Hill stability.
* Hill plot - Visualizes and assesses the Hill stability of orbits.
* Hill diameter - The maximum distance between two satellites in the same orbit plane to avoid collisions. |
| **Physics / Optics** | * Euler-Bernoulli beam theory - Modeling bending forces in beams using Euler's equations.
* Euler buckling - Failure mode of columns and beams under high compressive stresses.
* Euler angles - Used to describe 3D orientation of objects in physics and engineering.
* Euler turbine equations - Model the fluid flow and energy transfer in turbine machinery.
* Euler-Lagrange equations - Fundamental equations of motion in classical mechanics and field theory.
* Euler's laws of motion - Principles for rotation of rigid bodies about a fixed point.
* Euler hydrodynamic equations - Describe motion of an ideal fluid and form the basis of CFD.
* Euler model - Physical systems modeled using Euler's equations. |
| | * Lambert's absorption law - Exponential decrease of light passing through an absorbing medium.
* Lambert-Beer law - Relates absorption of light to properties of the material it passes through.
* Lambert's cosine law - Describes the intensity of light emitted from a surface as proportional to the cosine of the angle from the normal.
* Lambertian reflectance - The property of a surface to reflect incident light equally in all directions. |

| | |
|---|---|
| | * Lambertian emittance - The property of a surface to emit light equally in all directions based on its temperature.<br>* Lambert's absorption law - The exponential decrease in intensity of light passing through an absorbing medium.<br>* Lambertian scattering - Reflection/scattering of light such that brightness is the same regardless of viewing angle.<br>* Lambertian scattering models - Optical models that assume Lambertian scattering to simplify analysis.<br>* Lambertian photometric functions - Functions that model the luminance of Lambertian surfaces. |
| **Mathematics / Graph Theory** | * Euler's number (e) - The base of the natural logarithm, a fundamental mathematical constant.<br>* Euler's formula - Relates exponential functions and trigonometric functions using imaginary numbers.<br>* Euler's totient function ($\varphi$) - Gives the number of positive integers less than a number that are coprime to it.<br>* Euler's polynomial - A particular polynomial that arises in the study of elliptic integrals.<br>* Euler's identity - Equates e, i, $\pi$ and 1 in a beautiful relation. Considered one of the most remarkable formulas in mathematics.<br>* Euler angles - Three angles used to describe the orientation of a rigid body in 3D space.<br>* Euler integration - A numerical integration technique for differential equations based on Euler's methods.<br>* Euler mathematical model - Using Euler's equations and methods to model physical systems.<br>* Euler equation - Differential equations formulated by Euler to model real world systems like the motion of rigid bodies.<br>* Euler path - Traverses each edge of a graph exactly once.<br>* Euler circuit - Traverses each edge of a graph once, ending at the starting node.<br>* Euler characteristic - A topological invariant, important in geometry and topology. |
| | * Lambert W function - The inverse function of f(z) = z*e$\hat{z}$. Allows analytical solutions of certain equations.<br>* Lambert's continued fraction - A particular continued fraction convention studied by Lambert.<br>* Lambert series - An infinite series named after Lambert involving tangent function terms. |
| | * Hill differential equation - A differential equation modeling various physical systems like planetary motion. |
| **Cartography** | * Lambert conformal conic projection - Conformal map projection using one or two standard parallels.<br>* Lambert conformal conic map - Specific conformal conic map invented by Lambert.<br>* Lambert cylindrical equal-area projection - Cylindrical projection preserving area.<br>* Lambert azimuthal equal-area projection - Azimuthal projection preserving area. |

|  | * Hill shading - An illumination technique to create the illusion of depth on maps and 3D visualizations. |
| --- | --- |
| **Other Disciplines** | * Hill muscle (Physiology) - A posterior thigh muscle involved in hip and knee movements.<br>* Hill reaction (Biochemistry) - Light-dependent production of oxygen during photosynthesis.<br>* Hill criterion (Biochemistry) - Used to determine ligand binding affinity to proteins.<br>* Hill number (Ecology) - A mathematical measure of diversity used in ecology. |